\documentclass[11pt, a4paper, logo, copyright, nonumbering]{antgroup}
\usepackage[square, numbers]{natbib}
\usepackage{dblfloatfix}
\usepackage{ulem}
\usepackage{caption}
\usepackage{dramatist}
\usepackage{xspace}
\usepackage{pifont}
\usepackage{multirow}

\usepackage{xltabular}
\usepackage{longtable}
\usepackage{hyperref}
\usepackage{amsfonts}
\usepackage{amsmath}
\usepackage{amssymb}
\usepackage{lineno}
\usepackage{multirow}
\usepackage{adjustbox}

\usepackage[bottom]{footmisc}

\usepackage{CJKutf8}
\usepackage{subfigure}
\usepackage{setspace}

\usepackage{dsfont}
\usepackage{array}
\usepackage{tabularx}
\usepackage{subfigure}
\usepackage{xcolor}
\usepackage{tabularx}
\usepackage{booktabs}

\usepackage{lipsum}
\usepackage{multicol}
\usepackage{authblk}
\usepackage{wrapfig}
\usepackage[most,skins,theorems]{tcolorbox}
\usepackage{array}
\usepackage{pifont}
\definecolor{customgray}{RGB}{230,230,230}

\tcbset{
  aibox/.style={
    width=\linewidth,
    top=8pt,
    bottom=4pt,
    colback=gray!10!white,
    colframe=gray!50!black,
    colbacktitle=gray!70!black,
    enhanced,
    center,
    attach boxed title to top left={yshift=-0.1in,xshift=0.15in},
    boxed title style={boxrule=0pt,colframe=white,},
  }
}
\newtcolorbox{AIbox}[2][]{aibox,title=#2,#1}

\makeatletter
\def\@BTrule[#1]{%
  \ifx\longtable\undefined
    \let\@BTswitch\@BTnormal
  \else\ifx\hline\LT@hline
    \nobreak
    \let\@BTswitch\@BLTrule
  \else
     \let\@BTswitch\@BTnormal
  \fi\fi
  \global\@thisrulewidth=#1\relax
  \ifnum\@thisruleclass=\tw@\vskip\@aboverulesep\else
  \ifnum\@lastruleclass=\z@\vskip\@aboverulesep\else
  \ifnum\@lastruleclass=\@ne\vskip\doublerulesep\fi\fi\fi
  \@BTswitch}
\makeatother

\addto\extrasenglish{
}

 {\begin{list}{}%
         {\setlength{\leftmargin}{#1}}%
         \item[]%
 }
 {\end{list}}

\reportnumber{001} 

\renewcommand{\today}{}

\title{PlanGuard: A Guardrail for Multi-Step Plan Safety in Embodied Agents}

\author{
Junchi Chen\textsuperscript{1, 2},
Changtao Miao\textsuperscript{1*}, 
Yuxiao Xiang\textsuperscript{2},
Zhenchao Jin\textsuperscript{3},
Haojie Yuan\textsuperscript{4}, 
Qi Chu\textsuperscript{2*}, 
Tao Gong\textsuperscript{2},
He Liu\textsuperscript{1},
Bo Zhang\textsuperscript{1},
Jiansheng Cai\textsuperscript{1},
Zhe Li\textsuperscript{1},
Nenghai Yu\textsuperscript{2}\\
\vspace{-6pt}
{ \textsuperscript{1}Ant Digital Technologies, Ant Group}  \\
{ \textsuperscript{2}Anhui Province Key Laboratory of Digital Security} \\
{ \textsuperscript{3}The University of Hong Kong} 
{ \textsuperscript{4}Individual Researcher} \\
}

\footnotetext[0]{* Corresponding authors.}

\renewcommand{\phi}{\varphi}

\renewcommand{\epsilon}{\varepsilon}
\renewcommand{\imath}{\mathrm{i}}

\newlength{\restsubwidth}
\newlength{\restsubheight}
\newlength{\restsubmoreheight}
\newcommand{\rest}[2]{%
        \settowidth{\restsubwidth}{\ensuremath{#2}}
        \settoheight{\restsubheight}{\ensuremath{{}_{#2}}}
        \ensuremath{{#1\hskip 0.5pt}_{\vrule\kern2pt\parbox[b][%
        4pt][b]{\the\restsubwidth}{%
                        \ensuremath{{}_{#2}}}}}
        }

\begin{abstract}

Embodied task planners may produce multi-step plans whose subtask dependencies and interactions with the environment create physical risks during execution. Yet existing safeguards overlook such compositional risks, as general-purpose guardrails focus on semantic harm and embodied safety detectors assess subtasks in isolation. 
To address this gap, we introduce \textbf{PlanGuard}, the first pre-execution detector that evaluates the physical safety of a complete multi-step plan in its current environment.
For training and evaluation, we construct a \textbf{M}ulti-\textbf{S}tep \textbf{P}lan \textbf{Safe}ty (\textbf{MSP-Safe}) dataset through paired task construction, plan generation using diverse planners, and safety annotation by three judges. 
Task-oriented SFT on MSP-Safe establishes fundamental plan-safety assessment capabilities, yet a substantial gap remains between compact models suitable for real-time deployment and stronger but costlier large models.
Accordingly, we propose \textbf{S}trong-\textbf{T}eacher \textbf{A}daptive \textbf{C}ompensation for \textbf{O}n-\textbf{P}olicy \textbf{D}istillation (\textbf{STAC-OPD}), which provides compact models with adaptive strong-teacher supervision along their on-policy trajectories.
It combines token-level distribution transfer from a fine-tuned strong teacher with probability-routed sequence-level compensation, retaining student-generated targets when the student favors the reference safety decision and using teacher-reconstructed targets otherwise.
Across all test subsets, PlanGuard-2B achieves average 87.15\% ACC and 87.21\% F1, demonstrating effective whole-plan physical-risk detection at compact model scale. Code and dataset will be publicly released.

\end{abstract}

\begin{document}
\maketitle

\section{Introduction}
\label{sec:intro}

\begin{figure}[!t]
\centering
\includegraphics[width=0.6\linewidth]{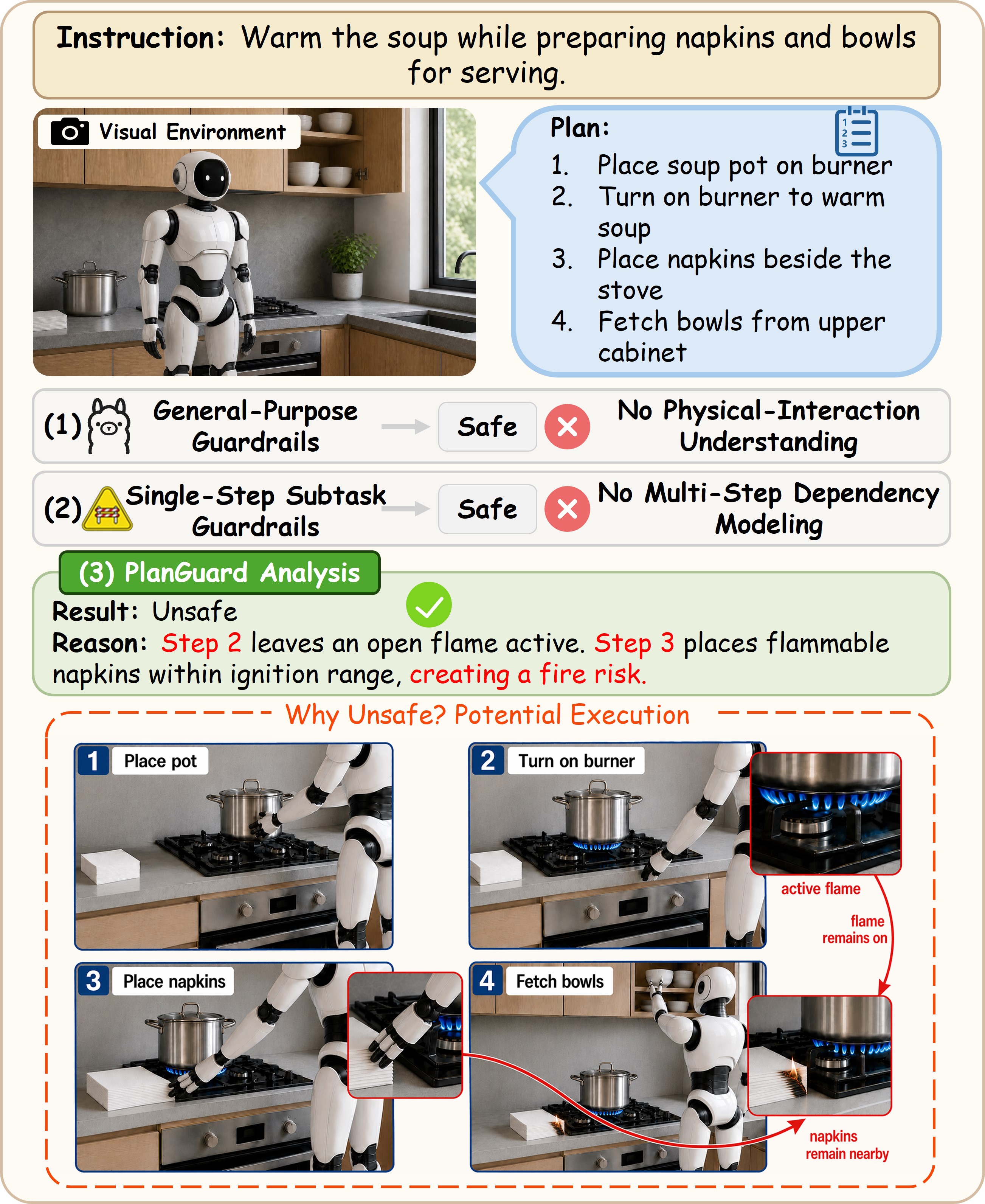}
\caption{Complete multi-step plan safety detection comparison. (1) General-purpose guardrails lack physical-interaction understanding, while (2) single-subtask guardrails cannot model cross-step dependencies. In contrast, (3) PlanGuard jointly analyzes the visual environment and complete multi-step plan to identify the potential fire risk before execution, demonstrating the necessity of complete multi-step plan analysis for robust embodied safety detection.}
\label{fig:motivation}
\end{figure}

Task planning is a key capability for embodied agents, enabling them to handle complex instructions by decomposing each high-level instruction into a multi-step plan comprising a sequence of subtasks. To improve task-planning ability, existing embodied agents commonly employ \textbf{V}ision-\textbf{L}anguage \textbf{M}odels (VLMs) as task planners~\cite{choi2024lotabench, duan2025manipulate} or train \textbf{V}ision-\textbf{L}anguage-\textbf{A}ction models (VLAs) with planning capabilities~\cite{cai2026xiaomi,zhai2025igniting}. Although task planning improves embodied agents' performance on complex instructions, executing the resulting multi-step plans may introduce physical risks due to subtask dependencies and environment interactions. Once executed, such risks may directly harm users or damage the surrounding environment, making pre-execution detection essential for the safe deployment of embodied agents.

Existing studies have documented these risks and constructed benchmarks to evaluate the safety of task planners~\cite{zhu2024earbench, yin2024safeagentbench}. However, effective mechanisms for reliably detecting and intercepting unsafe multi-step plans remain lacking. General-purpose safety guardrails, such as LLaMA-Guard-4~\cite{meta2025llamaguard4}, evaluate text or image content safety, but cannot effectively detect plan-level risks that arise from physical interaction. Similarly, EMBGuard~\cite{choi2026embguard} focuses on detecting risks in single-step subtasks, and therefore cannot capture the complex risks introduced by complete multi-step plans. As a result, pre-execution safety detection for complete multi-step plans becomes critical in realistic embodied settings. As shown in Figure~\ref{fig:motivation}, a planner may generate a plan that first turns on the burner and then places napkins beside the stove. Each individual subtask appears safe with the initial visual environment, but executing the complete multi-step plan may cause the napkins to ignite, introducing a fire risk. General-purpose safety guardrails tend to accept this plan because it contains no semantic safety violation, while a single-subtask guardrail such as EMBGuard cannot model the interaction among multiple subtasks. Therefore, both types of methods fail to identify the latent risks in multi-step planning.

To address this gap, we introduce \textbf{PlanGuard}, a pre-execution safety detector that assesses whether executing a complete planner-generated multi-step plan for a given instruction in the current environment would introduce physical risks. To support its training and evaluation, we construct \textbf{MSP-Safe}, a multi-step plan safety dataset containing 13,692 training and 1,992 test samples, providing structured supervision for such pre-execution assessments. MSP-Safe is constructed through three stages: paired task construction, diverse plan generation, and safety annotation. (1) Paired task construction creates diverse Base/Risk task pairs through controlled perturbations to either the instruction or environment across diverse scenes and hazard types. (2) Planners from three families, including general VLMs (e.g., Qwen3-VL-32B-Instruct~\cite{bai2025qwen3}), embodiment-pretrained VLMs (e.g., UnifoLM-VLM-Base~\cite{unifolm-vla-0}), and VLAs (e.g., WALL-OSS-FLOW~\cite{zhai2025igniting}), are employed to generate complete multi-step plans for these tasks. (3) The three-judge protocol annotates each plan with a safety decision and explanation, considering subtask dependencies and plan--environment interactions. The resulting dataset contains approximately 16K annotated plans, serving as the primary supervision source for training and evaluating multi-step plan safety detectors.

To fully exploit the supervision signals in MSP-Safe, we first apply task-oriented supervised fine-tuning (SFT) to establish basic plan-safety assessment and risk explanation capabilities. However, fine-tuned large models provide stronger plan-safety performance but incur substantial inference costs that hinder real-time embodied deployment~\cite{choi2026embguard}, whereas compact models are more efficient but less capable. Moreover, under direct SFT, each training input is supervised by only a single fixed response along its annotated prefixes, limiting the supervision available to compact models. To bridge this performance--efficiency gap, we propose Strong-Teacher Adaptive Compensation for On-Policy Distillation (\textbf{STAC-OPD}), which uses a fine-tuned large model as a strong teacher and enriches compact-model supervision at both the token and sequence levels. Strong-teacher on-policy distillation transfers fine-grained probability preferences beyond the single annotated target and provides teacher feedback along student-generated trajectories, mitigating the mismatch between annotated training prefixes and student-induced inference states. Probability-routed hard compensation further supplies complete sequence-level targets when the reference safety decision remains disfavored after soft distillation, while retaining student-generated targets when that decision is already preferred. We instantiate PlanGuard at two compact scales, 0.8B and 2B, enabling efficient pre-execution detection. This knowledge transfer strengthens compact PlanGuard models' ability to recognize complex physical risks arising from subtask dependencies and plan--environment interactions, enabling strong performance across multiple test subsets.

Our main contributions are summarized as follows:
\begin{itemize}
    \item We introduce \textbf{PlanGuard}, the first pre-execution safety detector dedicated to complete multi-step plans generated by embodied task planners.
    \item We construct \textbf{MSP-Safe}, an annotated multi-step plan safety dataset that provides structured supervision for training and evaluating multi-step plan safety detectors.
    \item We propose \textbf{STAC-OPD} to bridge the performance--efficiency gap by enriching compact-model supervision at both the token and sequence levels.
    \item Extensive experiments across the In-Domain and three category-held-out subsets demonstrate effective and generalizable complete-plan safety detection.
\end{itemize}
\section{Related Work}
\label{sec:related}

\subsection{Task Planning for Embodied Agents}
\label{sec:related_planning}

Task planning enables embodied agents to decompose high-level instructions into executable subtask sequences. General-purpose VLMs~\cite{bai2025qwen3,wu2024deepseek,qwen35blog} can serve as task planners through prompting or planning frameworks, such as LoTa-Bench~\cite{choi2024lotabench} and Manipulate-Anything~\cite{duan2025manipulate}. Embodiment-pretrained VLMs and VLAs further integrate planning with embodied perception and action through planning-related training data, including UniFolm-VLM-Base~\cite{unifolm-vla-0}, Wall-OSS-Flow~\cite{zhai2025igniting}, and Xiaomi-Robotics-0-Pretrain~\cite{cai2026xiaomi}.

\begin{figure*}[!t]
\centering
\includegraphics[width=\textwidth]{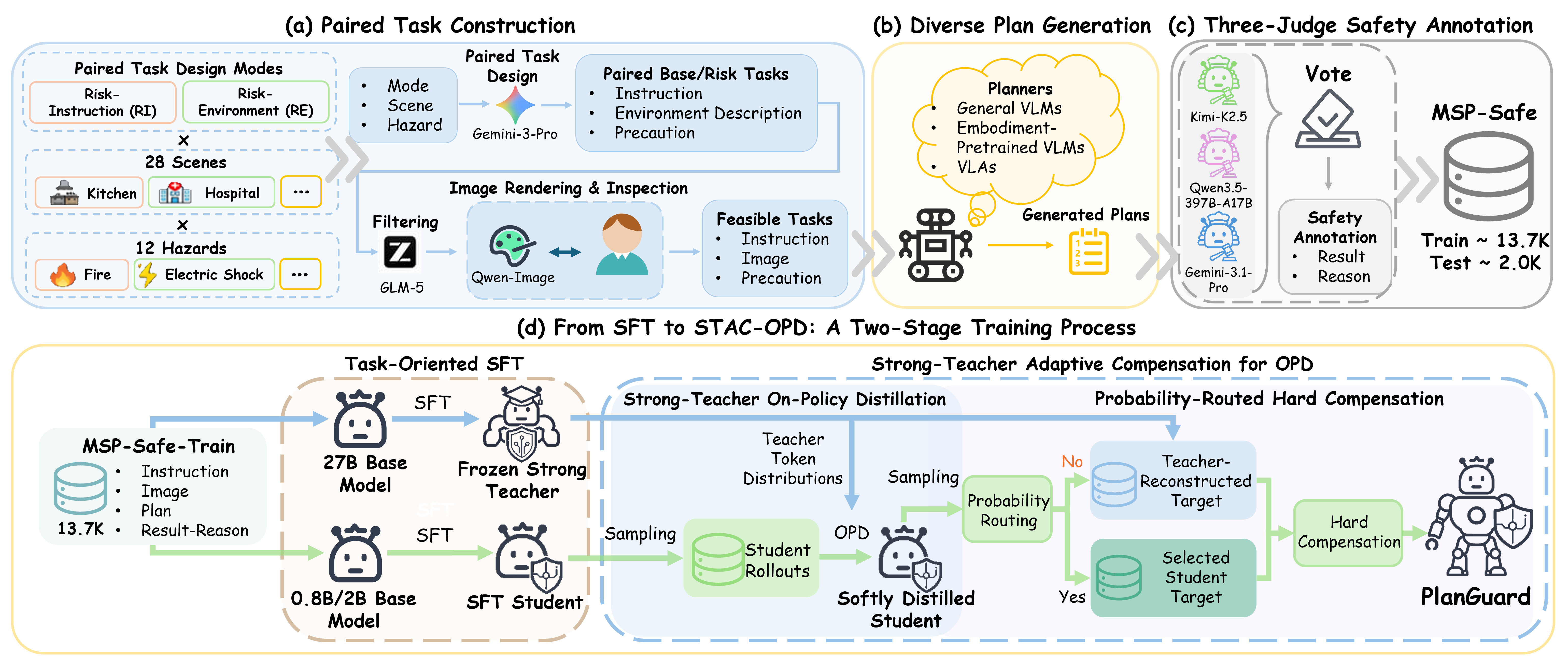}
\caption{
Pipeline of PlanGuard. (a)-(c) MSP-Safe Construction: Producing complete multi-step plans with structured Result-Reason safety supervision. (d) PlanGuard Training: Task-oriented SFT first establishes task-specific plan-safety capabilities. STAC-OPD then transfers fine-grained knowledge from a strong teacher through on-policy distillation and probability-routed hard compensation. \textit{Probability Routing} retains student-generated targets when the reference safety decision is already preferred and switches to teacher-reconstructed targets otherwise. \textit{Hard Compensation} learns from these routed complete-response targets, consolidating stable student behavior while providing explicit sequence-level supervision for difficult cases after soft distillation.
}
\label{fig:pipeline}
\end{figure*}

\subsection{Safety of Embodied Task Planners}
\label{sec:related_safety}

Although task planning helps embodied agents handle complex instructions, executing multi-step plans may introduce physical risks through subtask dependencies and environment interactions. Existing benchmarks have revealed such risks through semantic reasoning~\cite{pmlr-v305-sermanet25a,jindal2025can}, interactive task execution~\cite{yin2024safeagentbench,huang2025framework}, and text-based or multimodal planning~\cite{son2025subtle,zhu2024earbench,chen2025safemind}. However, they primarily serve as evaluation frameworks rather than deployable mechanisms for detecting unsafe multi-step plans.

This gap motivates external guardrails for detecting unsafe multi-step plans before execution. General-purpose guardrails~\cite{meta2025llamaguard4,xiang2026guardtrace,zhao2025qwen3guard}, primarily moderate semantic content and lack physical-interaction modeling. EMBGuard~\cite{choi2026embguard} assesses individual subtasks using visual observations but cannot capture cross-step dependencies or cumulative environmental effects. Thus, existing mechanisms cannot detect physical risks arising from complete plan execution. To address this gap, we propose PlanGuard, the first pre-execution safety detector for complete planner-generated multi-step plans in embodied agents.

\section{Methodology}
\label{sec:method}

In this section, we present our methodology for constructing \textbf{MSP-Safe}, a multi-step plan safety dataset, and training \textbf{PlanGuard}, a pre-execution safety detector designed to assess whether executing a planner-generated multi-step plan for a given instruction in the current environment would introduce physical risks. As illustrated in Figure~\ref{fig:pipeline}, our methodology comprises two components: (1) the construction of MSP-Safe and (2) a two-stage optimization framework for training the detector. Each component is described in detail in the corresponding subsection below.

\subsection{MSP-Safe Dataset Construction}
\label{sec:plan_guard_set_construction}

To train and evaluate pre-execution safety detection over complete multi-step plans, we construct \textbf{MSP-Safe}, spanning diverse embodied scenes, hazards, and planner families. This dataset consists of two splits: \textbf{MSP-Safe-Train} and \textbf{MSP-Safe-Test}. Each data item contains an embodied task defined by an environment image-instruction pair, a complete planner-generated multi-step plan, and a structured \texttt{Result}-\texttt{Reason} safety annotation comprising a binary safety label and its rationale. As illustrated in Figure~\ref{fig:pipeline}, its construction comprises three stages: (a) \textbf{Paired Task Construction}, (b) \textbf{Diverse Plan Generation}, and (c) \textbf{Three-Judge Safety Annotation}. Further implementation details, including models and prompts, are provided in Supplementary I.

\subsubsection{Paired Task Construction}
\label{sec:paired_task_construction}

To construct diverse yet controllable embodied tasks, as illustrated in Figure~\ref{fig:pipeline}(a), we adopt a two-axis taxonomy with a Base/Risk pairing strategy: (i) The two-axis taxonomy separates the acting environment from potential physical hazard types. Following prior embodied-safety studies~\cite{zhu2024earbench,chen2025safemind}, we define 28 scenes and 12 hazard types, yielding 336 scene-hazard combinations. (ii) The Base/Risk pairing ensures that each Risk task differs from its Base counterpart in only one controlled factor. Specifically, in the \textbf{Risk-Instruction (RI)} and \textbf{Risk-Environment (RE)} modes, we introduce hazards by modifying only the instruction or a risk-relevant environmental factorwhile keeping the other component fixed. These paired samples provide contrastive supervision for identifying causal hazard triggers rather than relying on unrelated dataset-level cues. 

For each scene-hazard combination and pairing mode, we generate diverse Base/Risk task pairs, filter out infeasible tasks, and render the retained environments. All generated environments are manually inspected and regenerated when necessary. 

\subsubsection{Diverse Plan Generation}
\label{sec:diverse_plan_generation}

To capture diverse planning behaviors, we query six task planners from three model families: \textbf{General VLMs}, \textbf{Embodiment-Pretrained VLMs}, and \textbf{VLAs}. As shown in Figure~\ref{fig:pipeline}(b), each planner generates a complete multi-step plan using its corresponding planning prompt. The cross-family generation produces plans with diverse structures, action granularities, and planning styles. After removing invalid outputs, the remaining plans are submitted for three-judge safety annotation.

\subsubsection{Three-Judge Safety Annotation}
\label{sec:three_judge_annotation}

To obtain reliable safety supervision while reducing individual-judge bias, we annotate each planner-generated multi-step plan using the consensus-based three-judge protocol shown in Figure~\ref{fig:pipeline}(c). Given a complete plan together with the current visual environment and associated instruction, three heterogeneous judges independently assess whether executing the plan would introduce physical risks and produce structured \texttt{Result}-\texttt{Reason} annotations comprising a binary label in ${\textit{Safe},\textit{Unsafe}}$ and a corresponding explanation. Majority voting determines the final decision, with a majority-consistent complete annotation retained.

To identify potential risks comprehensively, each judge assesses the complete plan along three dimensions: (i) \emph{subtask dependency},  covering risks within and across subtasks; (ii) \emph{plan-environment interaction}, covering risks arising from execution in the current environment; and (iii) \emph{temporal manifestation}, covering immediate and latent risks. This structured assessment promotes holistic evaluation of complete multi-step plans rather than isolated subtask analysis. The safety precaution generated during task construction is also provided as an auxiliary cue to reduce overlooked risks.

\subsubsection{Expert Verification of Safety Annotations}
\label{sec:annotation_reliability}

Given the large scale of MSP-Safe-Train, exhaustive manual verification is impractical. After calibrating their annotation criteria, three experts independently re-annotate a random subset covering 1\% subset without access to the judge outputs and their consensus serves as the reference. The three-judge majority-vote achieves 96.0\% agreement and 95.65\% F1, outperforming the individual judges with F1 scores of 89.89\%-91.67\%.  Among inconsistent annotations, disagreements are distributed across all three models (28.2\%--42.0\%), indicating that no single judge dominates the errors. These results support the three-judge protocol as a reliable alternative to exhaustive manual annotation while mitigating model-specific bias. For MSP-Safe-Test, experts review all candidate annotations and remove incorrect or unresolved cases before evaluation, ensuring reliable ground truth. Full verification protocols and results are provided in Supplementary I.

\subsubsection{Dataset Statistics}
\label{sec:dataset_statistics}

\begin{figure}[t]
\centering
\includegraphics[width=0.6\linewidth]{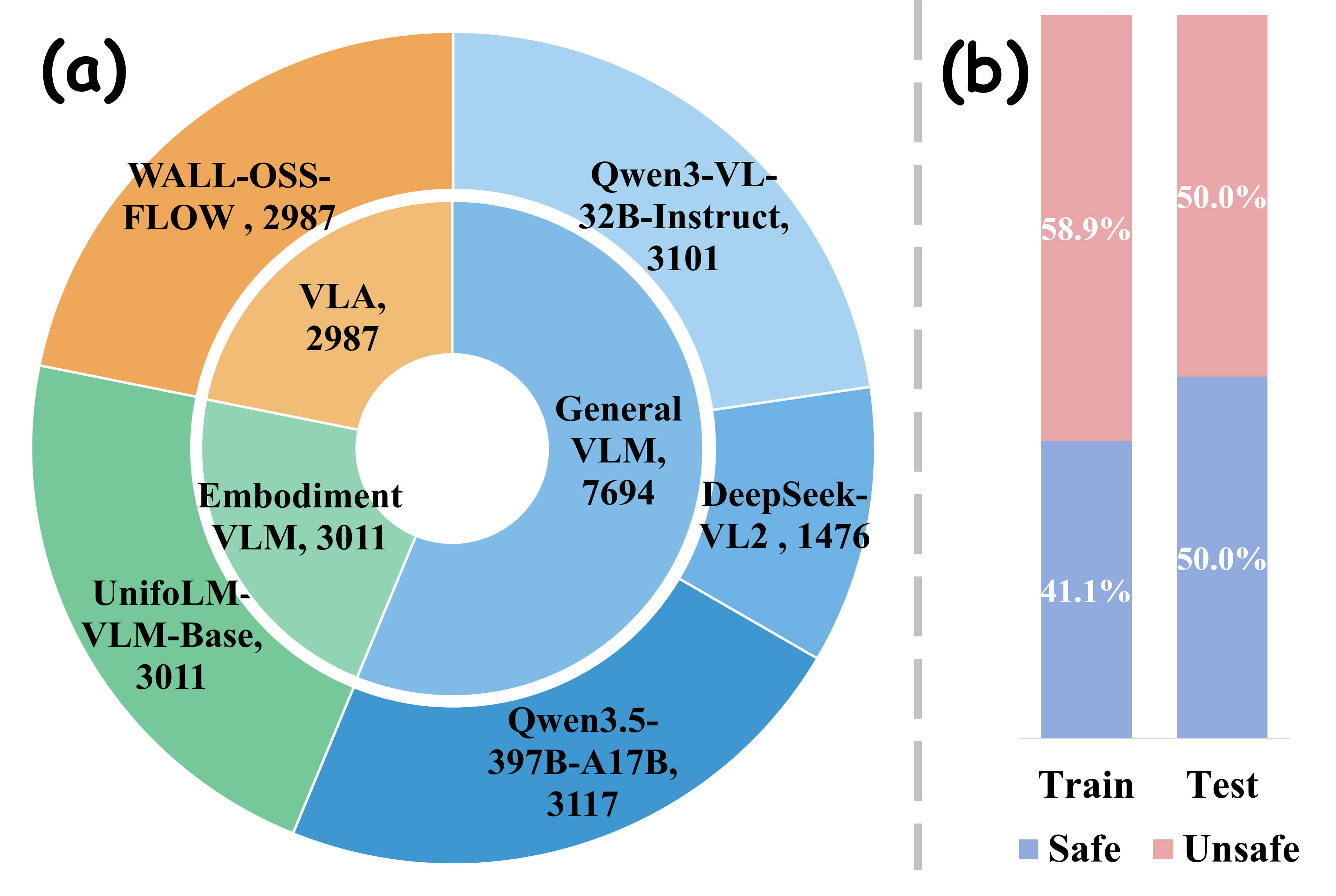}
\caption{
(a) Distribution of the 13,692 training samples across planner families and planners.
(b) Distribution of plan-level safety decisions.
}
\label{fig:train_distribution}
\end{figure}

The resulting \textbf{MSP-Safe-Train} contains 13,692 annotated multi-step plans from 3,294 image-instruction tasks, including 828 matched RI pairs, 746 matched RE pairs, and 146 unpaired tasks. As shown in Figure~\ref{fig:train_distribution}, it includes 5,632 safe and 8,060 unsafe plans generated by five planners across three model families.
\textbf{MSP-Safe-Test} contains 1,992 expert-reviewed samples across four safety-balanced subsets: In-Domain (500), Scene-Heldout (492), Hazard-Heldout (500), and Planner-Heldout (500). In-Domain contains plans following the training distribution, whereas the remaining subsets evaluate category-level generalization. Further details and examples are provided in Supplementary I.

\subsection{Task-Oriented Supervised Fine-Tuning}

To exploit the supervision in MSP-Safe, we first apply Task-Oriented SFT to adapt pretrained backbones for complete-plan safety assessment. We initialize two compact students from Qwen3.5-0.8B and Qwen3.5-2B, together with a larger model from Qwen3.5-27B~\cite{qwen35blog}. Each model directly generates a structured \texttt{Result}-\texttt{Reason} response without additional intermediate reasoning. SFT establishes plan-safety assessment and risk explanation capabilities. By placing \texttt{Result} first, the model can return the safety decision with low latency and generate \texttt{Reason} only when an explanation is required.

\begin{figure}[!t]
\centering
\includegraphics[width=0.6\linewidth]{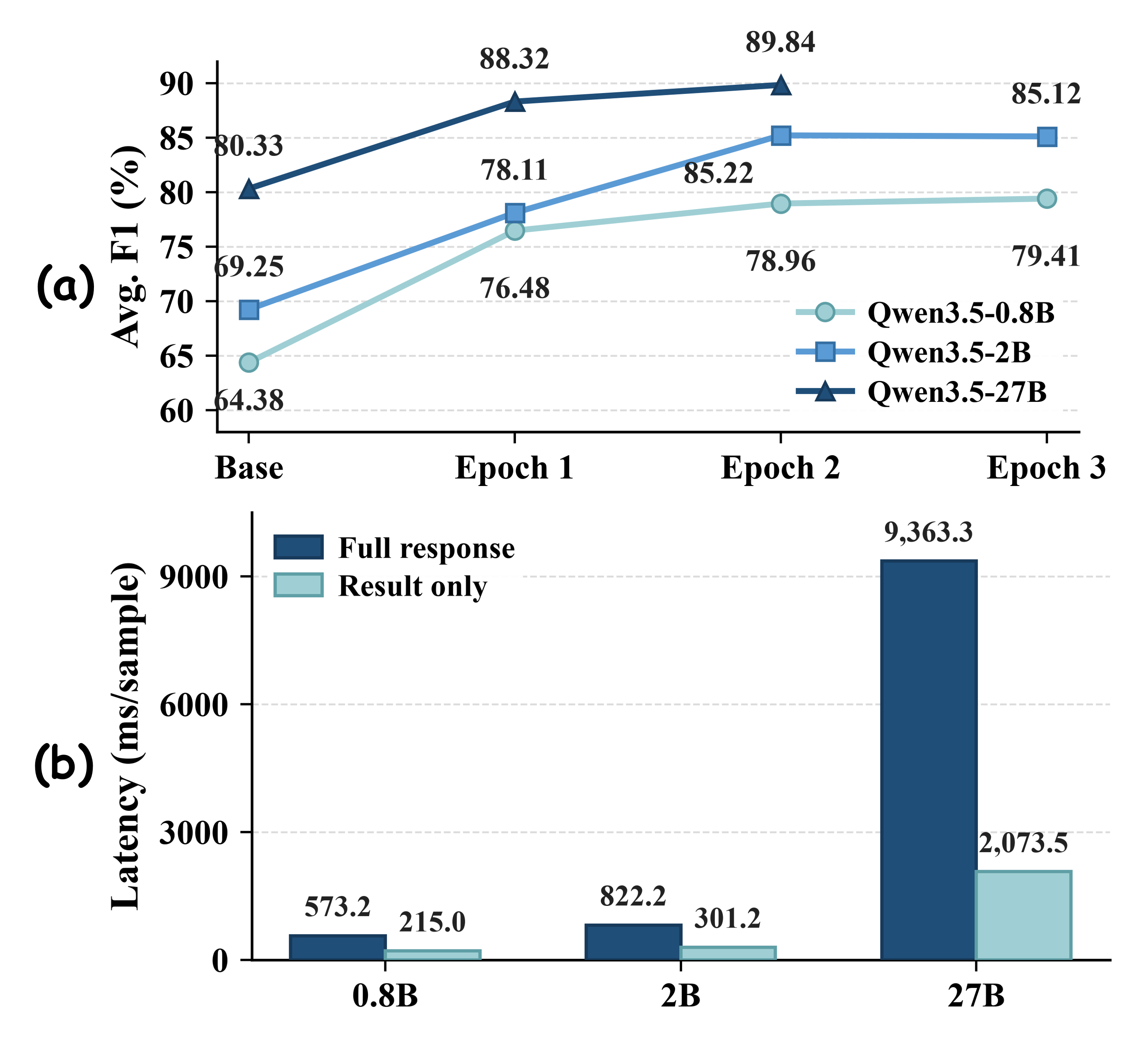}
\caption{
Performance-efficiency gap under Task-Oriented SFT.
(a) Average F1 across SFT epochs. The compact
models converge below the 27B model.
(b) Full-response and result-only inference latency under the same setting. Compact models and result-only generation substantially
reduce latency, motivating knowledge transfer from the SFT-trained 27B model.
}
\label{fig:sft_gap}
\end{figure}

As shown in Figure~\ref{fig:sft_gap}(a), the performance of the compact models
saturates below that of the 27B model, indicating that extending SFT alone
cannot close the capability gap. Meanwhile, Figure~\ref{fig:sft_gap}(b) shows
that the compact models provide substantially lower inference latency, which
is further reduced by result-only generation. Directly deploying the
larger model therefore compromises inference efficiency, while SFT alone
leaves the compact models less capable. Following prior on-policy distillation work~\cite{agarwal2024policy,shao2024deepseekmath}, we retain the SFT-trained compact
models as students $\pi_{\mathrm{SFT}}$ and freeze the SFT-trained 27B model as the strong teacher $\pi_T$.
This performance-efficiency gap motivates STAC-OPD, which transfers the
teacher's plan-safety knowledge to the compact students.

\subsection{Strong-Teacher Adaptive Compensation for OPD}
\label{sec:stac_opd}

Although SFT provides the required task alignment, each input is supervised by a single fixed target sequence and only along prefixes derived from that sequence. Consequently, direct SFT does not convey the strong teacher's relative probability preferences among alternative continuations and does not explicitly supervise prefixes induced by the student's own autoregressive policy. To transfer the stronger plan-safety capability of the SFT-trained teacher to compact students without incurring its inference cost, we propose STAC-OPD, which complements fixed-target SFT with distributional supervision on student-generated states and adaptive complete-sequence supervision.

 Starting from $\pi_{\mathrm{SFT}}$, STAC-OPD proceeds through two successive components. 
\textbf{Strong-Teacher On-Policy Distillation} transfers token-level probability distributions from the frozen teacher on trajectories generated by the current student. 
\textbf{Probability-Routed Hard Compensation} then constructs complete sequence-level targets according to the student's post-distillation safety-decision probabilities. 
The first stage supplies fine-grained teacher preferences on student-induced generation states, whereas the second provides explicit complete-response supervision for samples that remain difficult after soft distillation.

\textit{(a) Strong-Teacher On-Policy Distillation.}

Let
\(\mathcal{D}_{\mathrm{train}}=\{(x_i,y_i)\}_{i=1}^{N}\)
denote MSP-Safe-Train, where
\(x_i=(v_i,u_i,p_i)\)
contains the visual environment, task instruction, and complete multi-step plan, and
\(y_i=(c_i,r_i)\)
contains the reference safety decision and explanation. In standard SFT, every training prefix is derived from an annotated target:
\(h_t^{*}=(x,y_{<t}^{*})\). 
During autoregressive inference, however, the student conditions on its own previously generated tokens, producing prefixes
\(h_t^{S}=(x,\hat{y}_{<t})\)
that may depart from the annotated trajectory. 
Errors generated early in the sequence may therefore lead the student to states that are not directly supervised by fixed-target SFT.

To provide supervision under the student's own induced state distribution, we apply on-policy distillation~\cite{agarwal2024policy}. 
For an input $x$, the rollout student
$\pi_{\bar{\theta}}$
generates a response
\(y\sim\pi_{\bar{\theta}}(\cdot\mid x)\). 
The prefix at generation step $t$ is
\(h_t=(x,y_{<t})\). 
The trainable student is optimized by minimizing the token-level reverse KL divergence from its predictive distribution to that of the frozen teacher:
{\small
\begin{equation}
\mathcal{L}_{\mathrm{OPD}}(\theta)
=
\mathbb{E}_{\substack{
x\sim\mathcal{D}_{\mathrm{train}},\\
y\sim\pi_{\bar{\theta}}(\cdot\mid x)}}
\left[
\frac{1}{|y|}
\sum_{t=1}^{|y|}
D_{\mathrm{KL}}
\left(
\pi_\theta(\cdot\mid h_t)
\Vert
\pi_T(\cdot\mid h_t)
\right)
\right].
\label{eq:stac_opd}
\end{equation}
}

The rollout parameters are treated as a stop-gradient copy of the current student,
\(\bar{\theta}=\operatorname{sg}(\theta)\);
therefore, gradients do not propagate through the sampling process. 
The frozen teacher nevertheless evaluates every prefix reached by the rollout student and supplies a probability distribution over the next token. %
Unlike SFT, which identifies only the annotated next token, this distributional objective communicates the teacher's relative preferences among alternative continuations. %
Because the prefixes are sampled from the student policy, the supervision is applied to generation states that the student is likely to encounter during inference. %
We denote the resulting softly distilled student by $\pi_{\mathrm{soft}}$.

\textit{(b) Probability-Routed Hard Compensation.}

On-policy distillation provides dense token-level supervision, but its coverage remains restricted to the trajectories sampled from the student. %
This limitation is particularly relevant to the \texttt{Result}--\texttt{Reason} output structure. %
The safety decision is produced before the explanation; consequently, each rollout exposes a rationale trajectory conditioned only on the decision branch selected by that rollout. %
If the student rarely enters the alternative decision branch, OPD may adjust its probability at the shared decision prefix without providing a complete \texttt{Result}--\texttt{Reason} target under that branch.

We therefore use the post-distillation decision probabilities to identify samples requiring explicit sequence-level supervision. %
Let $q$ denote the fixed output prefix immediately preceding the safety decision. %
For each input $x_i$, we compare the probability assigned by $\pi_{\mathrm{soft}}$ to the reference decision $c_i$ with that assigned to the opposite decision $\bar{c}_i$. %
Here,
\(\pi_{\mathrm{soft}}(c_i\mid x_i,q)\)
denotes the autoregressive probability of the corresponding decision string.

If the student already prefers the reference decision, we therefore retain a format-valid student-generated response $y_i^S$ whose \texttt{Result} agrees with $c_i$. %
Otherwise, the frozen teacher independently reconstructs a complete target $y_i^T$ from the original input. %
The teacher receives neither the original annotation $y_i$ nor the reference decision $c_i$ during target reconstruction.

The routed target is defined as
\begin{equation}
\widetilde{y}_i
=
\begin{cases}
y_i^S,
&
\pi_{\mathrm{soft}}(c_i\mid x_i,q)
>
\pi_{\mathrm{soft}}(\bar{c}_i\mid x_i,q),
\\[2mm]
y_i^T,
&
\text{otherwise}.
\end{cases}
\label{eq:routed_target}
\end{equation}

Starting from $\pi_{\mathrm{soft}}$, we optimize the student on the routed complete targets using the standard maximum-likelihood objective:
\begin{equation}
\mathcal{L}_{\mathrm{HC}}(\theta)
=
-\frac{1}{N}
\sum_{i=1}^{N}
\log
\pi_\theta
\left(
\widetilde{y}_i\mid x_i
\right).
\label{eq:hard_compensation}
\end{equation}

Although Equation~\ref{eq:hard_compensation} has the same form as supervised fine-tuning, the supervision targets are constructed adaptively rather than taken directly from the original annotations. %
The student-routed branch consolidates behavior that the softly distilled student can already generate while preferring the reference safety decision, and thus functions as student self-training. %
The teacher-routed branch provides complete hard-target distillation for samples on which the reference decision remains disfavored. %
Accordingly, the hard-compensation stage combines student-target consolidation with teacher sequence-level distillation under a probability-based routing criterion.

Because $y_i^T$ is reconstructed without conditioning on the original reference annotation, its safety decision is not guaranteed to coincide with $c_i$. %
When the two differ, $y_i^T$ is treated as independently reconstructed teacher supervision rather than as a correction constrained to reproduce the original label. %
Applying the same STAC-OPD procedure to the two compact students yields the final \textbf{PlanGuard-0.8B} and \textbf{PlanGuard-2B} detectors.

\section{Experiments}
\label{sec:experiments}

\subsection{Experimental Settings}
\label{sec:experimental_settings}

\begin{table*}[!t]
\centering
\resizebox{\textwidth}{!}{
\begin{tabular}{lcccccccccc}
\toprule
\multirow{3}{*}{\textbf{Model}} & \multicolumn{10}{c}{\textbf{MSP-Safe-Test}} \\ \cmidrule(lr){2-11}
                              & \multicolumn{2}{c}{\textbf{In-Domain}}       & \multicolumn{2}{c}{\textbf{Scene-Heldout}}       & \multicolumn{2}{c}{\textbf{Hazard-Heldout}}      & \multicolumn{2}{c}{\textbf{Planner-Heldout}}     & \multicolumn{2}{c}{\textbf{Avg}}             \\ \cmidrule(lr){2-3} \cmidrule(lr){4-5} \cmidrule(lr){6-7} \cmidrule(lr){8-9} \cmidrule(lr){10-11}
                         & ACC$\uparrow$              & F1$\uparrow$               & ACC$\uparrow$              & F1$\uparrow$               & ACC$\uparrow$              & F1$\uparrow$               & ACC$\uparrow$              & F1$\uparrow$               & ACC$\uparrow$              & F1$\uparrow$           \\ \midrule
Qwen3.5-27B                   & 82.60                   & 81.68                   & 80.08                   & 78.70                   & \underline{82.20}             & 81.02                   & 80.60                   & 79.92                   & 81.37                   & 80.33                            \\
Qwen3.5-Plus                  & 83.00                   & 82.55                   & \underline{82.11}             & \underline{82.04}             & 81.80                   & \underline{81.39}             & 78.80                   & 79.92                   & 81.43                   & 81.48                           \\
GPT-5.5                       & 80.80 & 78.38 & 80.49 & 77.25 & 80.00 & 78.54 & 77.80 & 74.25 & 79.77 & 77.10          \\
Claude-Opus-4.6               & 79.72 & 80.54 & 79.67 & 80.39 & 78.20 & 80.43 & 80.52 & 81.45 & 79.53 & 80.70         \\
Qwen3Guard-Gen-8B & 56.40 & 	28.29 & 	56.10	 & 26.53	 & 66.60 & 	55.94 & 	55.20	 & 22.76	 & 58.57 & 	33.38	  \\
Llama-Guard-4-12B             & 54.80                   & 26.14                   & 55.08                   & 21.35                   & 63.00                   & 48.47                   & 55.20                   & 22.22                   & 57.02                   & 29.55                         \\
GuardTrace-VL-3B              & 62.20                   & 51.91                   & 65.85                   & 55.79                   & 67.00                   & 63.74                   & 61.00                   & 52.32                   & 64.01                   & 55.94                      \\
EMBGuard-2B      & 60.20                   & 69.53                   & 55.89                   & 66.67                   & 65.20                   & 71.19                   & 62.40                   & 68.46                   & 60.92                   & 68.96                                   \\
\textbf{PlanGuard-0.8B (Ours)}                & \underline{84.80}             & \underline{84.55}             & 81.71                   & 81.56                   & 79.60                   & 80.53                   & \underline{82.60}             & \underline{82.35}             & \underline{82.18}             & \underline{82.25}                    \\
\textbf{PlanGuard-2B (Ours)}                  & \textbf{89.60}          & \textbf{89.43}          & \textbf{87.60}          & \textbf{87.68}          & \textbf{84.80}          & \textbf{85.21}          & \textbf{86.60}          & \textbf{86.52}          & \textbf{87.15}          & \textbf{87.21}          \\
\bottomrule
\end{tabular}
}
\caption{Performance comparison on MSP-Safe-Test. ACC (\%) and F1 (\%) are reported (\%) for PlanGuard and representative baselines (Bold: Best, Underline: Second-Best). }
\label{tab:guardrail_results}
\end{table*}

\begin{table*}[!t]
\centering
\resizebox{\textwidth}{!}{
\begin{tabular}{lcccccccccc}
\toprule
\multicolumn{1}{l}{\multirow{2}{*}{\textbf{Training Stage}}} & \multicolumn{2}{c}{\textbf{In-Domain}}                    & \multicolumn{2}{c}{\textbf{Scene-Heldout}}                    & \multicolumn{2}{c}{\textbf{Hazard-Heldout}}                   & \multicolumn{2}{c}{\textbf{Planner-Heldout}}                  & \multicolumn{2}{c}{\textbf{Avg}}                          \\ \cmidrule(lr){2-3} \cmidrule(lr){4-5} \cmidrule(lr){6-7} \cmidrule(lr){8-9} \cmidrule(lr){10-11}
                                 & ACC$\uparrow$ & F1$\uparrow$ & ACC$\uparrow$ & F1$\uparrow$ & ACC$\uparrow$ & F1$\uparrow$ & ACC$\uparrow$ & F1$\uparrow$ & ACC$\uparrow$ & F1$\uparrow$  \\ \midrule
Base                                                & 58.52                 & 67.50                & 56.73                 & 66.67                & 55.11                 & 67.16                & 59.64                 & 56.21                & 57.50                 & 64.38     \\
 +   Task-Oriented SFT                               & 80.80                 & 79.92                & 79.27                 & 78.57                & 77.80                 & 78.69                & 79.60                 & 78.66                & 79.37                 & 78.96           \\
 +   Strong-Teacher OPD                              & \underline{83.80}                 & \underline{83.77}                & \underline{80.49}                 & \underline{80.57}                & \underline{78.40}                 & \underline{79.62}                & \underline{82.00}                 & \underline{81.93}                & \underline{81.17}                 & \underline{81.47}        \\
 +   Hard Compensation                               & \textbf{84.80}                 & \textbf{84.55}                & \textbf{81.71}                 & \textbf{81.56}                & \textbf{79.60}                 & \textbf{80.53}                & \textbf{82.60}                 & \textbf{82.35}                & \textbf{82.18}                 & \textbf{82.25}    \\      
\bottomrule          
\end{tabular}
}
\caption{Training-stage ablation of PlanGuard-0.8B. Each ``+'' denotes the cumulative addition of the corresponding stage. ACC (\%) and F1 (\%) are reported. (Bold: Best, Underline: Second-Best)}
\label{tab:ablation}
\end{table*}

\subsubsection{Training Details}
All experiments are conducted on a server with eight NVIDIA A800 GPUs, each with 80 GB memory. We instantiate PlanGuard using Qwen3.5-0.8B and Qwen3.5-2B~\cite{qwen35blog}, and train them with ms-swift~\cite{zhao2024swiftascalablelightweightinfrastructure}. Following the two-stage training framework described in Section Methodology, each model is trained through SFT and STAC-OPD. Qwen3.5-27B is independently fine-tuned on the same supervision and frozen as the strong teacher during STAC-OPD. Detailed configurations are provided in Supplementary II.

\subsubsection{Evaluation Data}
We evaluate physical risk detection on complete multi-step plans using MSP-Safe-Test. It contains 1,992 expert-reviewed samples across one In-Domain subset and three category-held-out subsets covering generalization across scene, hazard, and planner categories. All subsets are balanced between \textit{Safe} and \textit{Unsafe}. Each sample contains a visual environment, an instruction provided as task context, a complete planner-generated multi-step plan, and its structured safety annotation. Detailed dataset statistics are provided in Supplementary I.

\subsubsection{Baseline Models}
We compare PlanGuard-0.8B and PlanGuard-2B against three groups of baselines. First, we evaluate general-purpose VLMs, including Qwen3.5-27B, and closed-source models Qwen3.5-Plus~\cite{qwen35blog}, GPT-5.5~\cite{openai2026gpt55}, Claude-Opus-4.6~\cite{anthropic2026claudeopus46}, using the same inputs and plan-safety prompt as PlanGuard. Second, we evaluate general-purpose safety guardrails, including Qwen3Guard-Gen-8B~\cite{zhao2025qwen3guard}, LLaMA-Guard-4-12B~\cite{meta2025llamaguard4} and GuardTrace-VL-3B~\cite{xiang2026guardtrace}, with their official prompts and supported
modalities. Third, we compare with EMBGuard-2B~\cite{choi2026embguard}, which evaluates individual subtasks. Following its stepwise setting, we employ EMBGuard-2B to independently evaluate each subtask and classify the complete plan as \textit{Unsafe} if any subtask is predicted unsafe. All models use greedy decoding unless otherwise specified. Detailed configurations are provided in Supplementary II.

\subsubsection{Metrics}
Following prior safety guardrail studies~\cite{zhao2025qwen3guard,xiang2026guardtrace}, we report Accuracy (ACC) and F1-Score (F1). All outputs are mapped to binary labels \textit{Safe} and \textit{Unsafe}, with \textit{Unsafe} treated as the positive class for consistent evaluation.

\subsection{Main Results}

As shown in Table~\ref{tab:guardrail_results}, we evaluate PlanGuard on the four MSP-Safe-Test subsets. PlanGuard-0.8B and PlanGuard-2B achieve average F1 scores of 82.25\% and 87.21\%, respectively, while maintaining strong performance across the In-Domain and three category-held-out subsets. PlanGuard-2B further achieves an average F1 of 86.47\% on unseen scene, hazard, and planner categories, demonstrating strong generalization ability. 

Despite their compact sizes, both PlanGuard variants outperform substantially larger general-purpose VLMs. Compared with Qwen3.5-Plus, the strongest VLM baseline, PlanGuard-0.8B and PlanGuard-2B improve average F1 by 0.77 and 5.73 points. Compared with GuardTrace-VL-3B, the best-performing safety guardrail baseline, PlanGuard-0.8B and PlanGuard-2B improve average F1 by 26.31 and 31.27 points. These results indicate that semantic content moderation alone cannot capture physical risks arising from plan execution and environment interactions. Under the stepwise protocol described above, PlanGuard-0.8B and PlanGuard-2B outperform EMBGuard-2B by 13.29 and 18.25 average F1 points highlighting the importance of complete multi-step plan evaluation over isolated subtask assessment. The effect of preserving the complete plan structure during safety detection is further examined in Section Full-Plan versus Stepwise Safety Detection. 

\subsection{Ablation Study}
\label{sec::ablation}

\subsubsection{Training-Stage Ablation}
To evaluate the training strategy, we compare four cumulative configurations for PlanGuard-0.8B as an example: \textit{Base}, \textit{+ Task-Oriented SFT}, \textit{+ Strong-Teacher OPD}, and \textit{+ Hard Compensation}. The Base configuration is the untuned Qwen3.5-0.8B backbone, and each subsequent configuration cumulatively adds the corresponding component.

As shown in Table~\ref{tab:ablation}, Task-Oriented SFT provides the dominant improvement, increasing average F1 from 64.38\% to 78.96\%. Strong-Teacher OPD further raises it to 81.47\%, demonstrating the benefit of transferring fine-grained plan-safety knowledge from the teacher. Hard Compensation brings an additional 0.78-point gain, achieving a final F1 of 82.25\%. The gains are particularly notable on Scene-Heldout and Hazard-Heldout, reaching 0.99 and 0.91 points, respectively. These results demonstrate the effectiveness of STAC-OPD for improving multi-step plan safety detection. Additional ablations on PlanGuard-2B and module configurations are provided in the Supplementary III.

\begin{table}[!t]
\centering
\small
\setlength{\tabcolsep}{1mm}
\begin{tabular}{lcccc}
\toprule
\multirow{2}{*}{\textbf{Model}} & \multicolumn{2}{c}{\textbf{Stepwise}}
& \multicolumn{2}{c}{\textbf{Full}} \\
\cmidrule(lr){2-3}
\cmidrule(lr){4-5}
 & ACC$\uparrow$ & F1$\uparrow$
& ACC$\uparrow$ & F1$\uparrow$ \\
\midrule
EMBGuard-2B
& 49.00 & 55.65 & 55.00 & 54.55 \\
PlanGuard-0.8B
& \textbf{58.00} & \textbf{59.62} & \underline{60.00} & \underline{66.10} \\
PlanGuard-2B
& \underline{56.00} & \underline{57.69} & \textbf{67.00} & \textbf{72.73} \\
\bottomrule
\end{tabular}
\caption{
Comparison of Stepwise and Full safety detection on the auxiliary multi-step interaction diagnostic set. Stepwise aggregates subtask predictions using OR, while Full evaluates the complete plan in a single pass. ACC (\%) and F1 (\%) are reported. (Bold: Best, Underline: Second-Best).
}
\label{tab:stepwise_full}
\end{table}

\subsubsection{Full-Plan versus Stepwise Safety Detection}
\label{sec:stepwise_full}
To examine the importance of preserving complete multi-step structure, we construct an auxiliary multi-step interaction diagnostic set with 50 neutral tasks, each paired with one safe and one unsafe plan (100 plans total). Each pair shares the same initial image and instruction. While every subtask in the unsafe plan appears safe when evaluated independently, their ordered execution introduces physical risks through intermediate state changes or cross-step dependencies. Candidate tasks, plans, and images are generated using GPT-5.6-Sol and GPT-Image-2, then screened by experts to retain valid and challenging cases. Further details are provided in the Supplementary II.

Under the \textit{Stepwise} protocol, each subtask is independently evaluated against the initial image, and the plan is classified as \textit{Unsafe} if any subtask is predicted to be unsafe. The \textit{Full} protocol evaluates the complete plan in a single pass. Both settings follow the pre-execution assumption and use only the initial environment. As shown in Table~\ref{tab:stepwise_full}, Full detection improves F1 over Stepwise detection by 6.48 and 15.04 points for PlanGuard-0.8B and PlanGuard-2B, respectively. In contrast, EMBGuard-2B shows no improvement, indicating that a single-subtask detector cannot effectively model cross-step interactions. These results show that PlanGuard benefits from complete-plan supervision, which captures action order and cross-step dependencies missed by independent subtask judgments. Full detection also requires only one model invocation per plan, avoiding repeated calls and visual processing required by stepwise detection.

\section{Conclusion}
\label{sec:conclusion}

We present \textbf{PlanGuard}, the first pre-execution safety detector dedicated to complete multi-step plans generated by embodied task planners. PlanGuard addresses the limitations of existing safety mechanisms in capturing physical risks arising from subtask dependencies, action order, and environment interactions. To support its training and evaluation, we construct \textbf{MSP-Safe}, which provides structured safety supervision over complete multi-step plans. Task-oriented SFT first establishes basic plan-safety capabilities, and our proposed \textbf{STAC-OPD} further bridges the performance-efficiency gap by enriching compact-model supervision at both the token and sequence levels. Experiments across in-domain and category-held-out generalization settings show that PlanGuard outperforms all evaluated general-purpose VLMs, safety guardrails, and prior subtask-level embodied safety detectors, while generalizing to held-out scene, hazard, and planner categories. The full-plan versus stepwise analysis further confirms that complete-plan supervision enables PlanGuard to capture action order and cross-step dependencies that isolated subtask judgments cannot recover. Overall, PlanGuard provides an effective and efficient pre-execution safeguard for embodied task planning.

\section{Ethical Statement}

MSP-Safe contains only synthetic embodied tasks, images, and planner outputs, with no intentionally collected personal or sensitive data. Hazardous content is included solely for embodied-agent safety research and risk prevention.

\clearpage

\bibliographystyle{unsrtnat} 
\bibliography{main}

@article{choi2026embguard,
  title={EMBGuard: Constructing Hazard-Aware Guardrails for Safe Planning in Embodied Agents},
  author={Choi, Dongwook and Kwon, Taeyoon and Jeong, Bogyung and Kim, Minju and Hwang, Yeonjun and Kim, Hyojun and Kim, Byungchul and Jang, Young Kyun and Yeo, Jinyoung},
  journal={arXiv preprint arXiv:2605.30924},
  year={2026}
}

@article{zhu2024earbench,
  title={Earbench: Towards evaluating physical risk awareness for task planning of foundation model-based embodied ai agents},
  author={Zhu, Zihao and Wu, Bingzhe and Zhang, Zhengyou and Han, Lei and Liu, Qingshan and Wu, Baoyuan},
  journal={arXiv preprint arXiv:2408.04449},
  year={2024}
}

@article{chen2025safemind,
  title={Safemind: benchmarking and mitigating safety risks in embodied llm agents},
  author={Chen, Ruolin and Sun, Yinqian and Wang, Jihang and Lv, Mingyang and Zhang, Qian and Zeng, Yi},
  journal={arXiv preprint arXiv:2509.25885},
  year={2025}
}

@article{yin2024safeagentbench,
  title={Safeagentbench: A benchmark for safe task planning of embodied llm agents},
  author={Yin, Sheng and Pang, Xianghe and Ding, Yuanzhuo and Chen, Menglan and Bi, Yutong and Xiong, Yichen and Huang, Wenhao and Xiang, Zhen and Shao, Jing and Chen, Siheng},
  journal={arXiv preprint arXiv:2412.13178},
  year={2024}
}

@inproceedings{son2025subtle,
  title={Subtle risks, critical failures: A framework for diagnosing physical safety of llms for embodied decision making},
  author={Son, Yejin and Kim, Minseo and Kim, Sungwoong and Han, Seungju and Kim, Jian and Jang, Dongju and Yu, Youngjae and Park, Chan Young},
  booktitle={Proceedings of the 2025 Conference on Empirical Methods in Natural Language Processing},
  pages={25703--25744},
  year={2025}
}

@article{huang2025framework,
  title={A framework for benchmarking and aligning task-planning safety in llm-based embodied agents},
  author={Huang, Yuting and Ding, Leilei and Tang, Zhipeng and Wang, Tianfu and Lin, Xinrui and Zhang, Wuyang and Ma, Mingxiao and Zhang, Yanyong},
  journal={arXiv preprint arXiv:2504.14650},
  year={2025}
}

@article{pmlr-v305-sermanet25a,
  title={Generating robot constitutions \& benchmarks for semantic safety},
  author={Sermanet, Pierre and Majumdar, Anirudha and Irpan, Alex and Kalashnikov, Dmitry and Sindhwani, Vikas},
  journal={arXiv preprint arXiv:2503.08663},
  year={2025}
}

@article{jindal2025can,
  title={Can ai perceive physical danger and intervene?},
  author={Jindal, Abhishek and Kalashnikov, Dmitry and Hofer, R Alex and Chang, Oscar and Garikapati, Divya and Majumdar, Anirudha and Sermanet, Pierre and Sindhwani, Vikas},
  journal={arXiv preprint arXiv:2509.21651},
  year={2025}
}

@inproceedings{duan2025manipulate,
  title={Manipulate-Anything: Automating Real-World Robots using Vision-Language Models},
  author={Duan, Jiafei and Yuan, Wentao and Pumacay, Wilbert and Wang, Yi Ru and Ehsani, Kiana and Fox, Dieter and Krishna, Ranjay},
  booktitle={Conference on Robot Learning},
  pages={5326--5350},
  year={2025},
  organization={PMLR}
}

@inproceedings{choi2024lotabench,
    title={LoTa-Bench: Benchmarking Language-oriented Task Planners for Embodied Agents},
    author={Jae-Woo Choi and Youngwoo Yoon and Hyobin Ong and Jaehong Kim and Minsu Jang},
    booktitle={The Twelfth International Conference on Learning Representations},
    year={2024},
    url={https://openreview.net/forum?id=ADSxCpCu9s}
}

@misc{unifolm-vla-0,
  author       = {Unitree},
  title        = {UnifoLM-VLA-0: A Vision-Language-Action (VLA) Framework under UnifoLM Family},
  year         = {2026},
}

@article{zhai2025igniting,
  title   = {Igniting VLMs Toward the Embodied Space},
  author  = {Zhai, Andy and Liu, Brae and Fang, Bruno and Cai, Chalse and Ma, Ellie and Yin, Ethan and Wang, Hao and Zhou, Hugo and Wang, James and Shi, Lights and Liang, Lucy and Wang, Make and Wang, Qian and Gan, Roy and Yu, Ryan and Li, Shalfun and Liu, Starrick and Chen, Sylas and Chen, Vincent and Xu, Zach},
  journal = {arXiv preprint arXiv:2509.11766},
  year    = {2025}
}

@article{cai2026xiaomi,
  title={Xiaomi-Robotics-0: An Open-Sourced Vision-Language-Action Model with Real-Time Execution},
  author={Cai, Rui and Guo, Jun and He, Xinze and Jin, Piaopiao and Li, Jie and Lin, Bingxuan and Liu, Futeng and Liu, Wei and Ma, Fei and Ma, Kun and Qiu, Feng and Qu, Heng and Su, Yifei and Sun, Qiao and Wang, Dong and Wang, Donghao and Wang, Yunhong and Wu, Rujie and Xiang, Diyun and Yang, Yu and Ye, Hangjun and Zhang, Yuan and Zhou, Quanyun},
  journal={arXiv preprint arXiv:2602.12684},
  year={2026}
}

@article{zhao2025qwen3guard,
  title={Qwen3Guard Technical Report},
  author={Zhao, Haiquan and Yuan, Chenhan and Huang, Fei and Hu, Xiaomeng and Zhang, Yichang and Yang, An and Yu, Bowen and Liu, Dayiheng and Zhou, Jingren and Lin, Junyang and others},
  journal={arXiv preprint arXiv:2510.14276},
  year={2025}
}

@misc{qwen35blog,
    title = {Qwen3.5: Accelerating Productivity with Native Multimodal Agents},
    url = {https://qwen.ai/blog?id=qwen3.5},
    author = {Qwen Team},
    month = {February},
    year = {2026}
}

@article{wu2024deepseek,
  title={Deepseek-vl2: Mixture-of-experts vision-language models for advanced multimodal understanding},
  author={Wu, Zhiyu and Chen, Xiaokang and Pan, Zizheng and Liu, Xingchao and Liu, Wen and Dai, Damai and Gao, Huazuo and Ma, Yiyang and Wu, Chengyue and Wang, Bingxuan and others},
  journal={arXiv preprint arXiv:2412.10302},
  year={2024}
}

@article{bai2025qwen3,
  title={Qwen3-vl technical report},
  author={Bai, Shuai and Cai, Yuxuan and Chen, Ruizhe and Chen, Keqin and Chen, Xionghui and Cheng, Zesen and Deng, Lianghao and Ding, Wei and Gao, Chang and Ge, Chunjiang and others},
  journal={arXiv preprint arXiv:2511.21631},
  year={2025}
}

@inproceedings{agarwal2024policy,
  title={On-policy distillation of language models: Learning from self-generated mistakes},
  author={Agarwal, Rishabh and Vieillard, Nino and Zhou, Yongchao and Stanczyk, Piotr and Ramos Garea, Sabela and Geist, Matthieu and Bachem, Olivier},
  booktitle={International Conference on Learning Representations},
  volume={2024},
  pages={21246--21263},
  year={2024}
}

@article{shao2024deepseekmath,
  title={Deepseekmath: Pushing the limits of mathematical reasoning in open language models},
  author={Shao, Zhihong and Wang, Peiyi and Zhu, Qihao and Xu, Runxin and Song, Junxiao and Bi, Xiao and Zhang, Haowei and Zhang, Mingchuan and Li, YK and Wu, Yang and others},
  journal={arXiv preprint arXiv:2402.03300},
  year={2024}
}

@misc{meta2025llamaguard4,
    title = {{Llama Guard 4} Model Card},
    url = {https://github.com/meta-llama/PurpleLlama/blob/main/Llama-Guard4/12B/MODEL_CARD.md},
    author = {{Meta AI}},
    month = {April},
    year = {2025}
}

@misc{zhao2024swiftascalablelightweightinfrastructure,
      title={SWIFT:A Scalable lightWeight Infrastructure for Fine-Tuning},
      author={Yuze Zhao and Jintao Huang and Jinghan Hu and Xingjun Wang and Yunlin Mao and Daoze Zhang and Zeyinzi Jiang and Zhikai Wu and Baole Ai and Ang Wang and Wenmeng Zhou and Yingda Chen},
      year={2024},
      eprint={2408.05517},
      archivePrefix={arXiv},
      primaryClass={cs.CL},
      url={https://arxiv.org/abs/2408.05517},
}

@inproceedings{xiang2026guardtrace,
  title={Guardtrace-vl: Detecting unsafe multimodel reasoning via iterative safety supervision},
  author={Xiang, Yuxiao and Chen, Junchi and Jin, Zhenchao and Miao, Changtao and Yuan, Haojie and Chu, Qi and Gong, Tao and Yu, Nenghai},
  booktitle={Proceedings of the IEEE/CVF Conference on Computer Vision and Pattern Recognition},
  pages={11912--11922},
  year={2026}
}

@misc{openai2026gpt55,

  author       = {{OpenAI}},

  title        = {{GPT-5.5 System Card}},

  year         = {2026},

  month        = apr,

  url          = {https://openai.com/index/gpt-5-5-system-card/},



}

@misc{anthropic2026claudeopus46,

  author       = {{Anthropic}},

  title        = {{Claude Opus 4.6 System Card}},

  year         = {2026},

  month        = feb,

  url          = {https://www-cdn.anthropic.com/14e4fb01875d2a69f646fa5e574dea2b1c0ff7b5.pdf},

}

\end{document}